\documentclass[sigconf,preprint,authorversion]{acmart} %for arXiv
\usepackage[utf8]{inputenc}
\usepackage{bbold}
\usepackage{amsmath,mathtools}
\usepackage[inline]{enumitem}
\usepackage{flushend}
\usepackage[detect-weight=true]{siunitx}
\usepackage{url,hyperref,cleveref}
\usepackage[linesnumbered]{algorithm2e}
\usepackage{subcaption}
\crefname{algocf}{algorithm}{algorithms}
\Crefname{algocf}{Algorithm}{Algorithms}
\usepackage{color,colortbl,etoolbox,xcolor}
\robustify\bfseries
\usepackage{csquotes}
\usepackage[normalem]{ulem}

\usepackage[export]{adjustbox}

\AtBeginDocument{%
  \providecommand\BibTeX{{%
    \normalfont B\kern-0.5em{\scshape i\kern-0.25em b}\kern-0.8em\TeX}}}

\copyrightyear{2022}
\acmYear{2022}
\setcopyright{acmcopyright}\acmConference[GECCO '22 Companion]{Genetic and Evolutionary Computation Conference Companion}{July 9--13, 2022}{Boston, MA, USA} \acmBooktitle{Genetic and Evolutionary Computation Conference Companion (GECCO '22 Companion), July 9--13, 2022, Boston, MA, USA}
\acmPrice{15.00}
\acmDOI{10.1145/3520304.3533983}
\acmISBN{978-1-4503-9268-6/22/07}

\begin{document}

%%
%% The "title" command has an optional parameter,
%% allowing the author to define a "short title" to be used in page headers.
\title{Reinforcement learning based adaptive metaheuristics}

%%
%% The "author" command and its associated commands are used to define
%% the authors and their affiliations.
%% Of note is the shared affiliation of the first two authors, and the
%% "authornote" and "authornotemark" commands
%% used to denote shared contribution to the research.

\author{Michele Tessari}
%\authornote{}
%\authornotemark[1]
\affiliation{%
%Department of Information Engineering and Computer Science, 
  \institution{University of Trento}
  \city{Trento}%Povo
  \country{Italy}
  %\streetaddress{Via Sommarive 9}
  %\postcode{38123}
}
\orcid{0000-0003-4398-4209}
\email{michele.tessari@studenti.unitn.it}

\author{Giovanni Iacca}
\affiliation{%
%Department of Information Engineering and Computer Science, 
  \institution{University of Trento}
  \city{Trento}%Povo
  \country{Italy}
  %\streetaddress{Via Sommarive 9}
  %\postcode{38123}
}
\orcid{0000-0001-9723-1830}
\email{giovanni.iacca@unitn.it}

%%
%% By default, the full list of authors will be used in the page
%% headers. Often, this list is too long, and will overlap
%% other information printed in the page headers. This command allows
%% the author to define a more concise list
%% of authors' names for this purpose.
\renewcommand{\shortauthors}{Tessari and Iacca}

\begin{abstract}
Parameter adaptation, that is the capability to automatically adjust an algorithm's hyperparameters depending on the problem being faced, is one of the main trends in evolutionary computation applied to numerical optimization. While several handcrafted adaptation policies have been proposed over the years to address this problem, only few attempts have been done so far at applying machine learning to learn such policies. Here, we introduce a general-purpose framework for performing parameter adaptation in continuous-domain metaheuristics based on state-of-the-art reinforcement learning algorithms. We demonstrate the applicability of this framework on two algorithms, namely Covariance Matrix Adaptation Evolution Strategies (CMA-ES) and Differential Evolution (DE), for which we learn, respectively, adaptation policies for the step-size (for CMA-ES), and the scale factor and crossover rate (for DE). We train these policies on a set of 46 benchmark functions at different dimensionalities, with various inputs to the policies, in two settings: one policy per function, and one global policy for all functions. Compared, respectively, to the Cumulative Step-size Adaptation (CSA) policy and to two well-known adaptive DE variants (iDE and jDE), our policies are able to produce competitive results in the majority of cases, especially in the case of DE.
\end{abstract}

%%
%% The code below is generated by the tool at http://dl.acm.org/ccs.cfm.
%% Please copy and paste the code instead of the example below.
%%


%%
%% Keywords. The author(s) should pick words that accurately describe
%% the work being presented. Separate the keywords with commas.

%% A "teaser" image appears between the author and affiliation
%% information and the body of the document, and typically spans the
%% page.
%\begin{teaserfigure}
%  \includegraphics[width=\textwidth]{sampleteaser}
%  \caption{Seattle Mariners at Spring Training, 2010.}
%  \Description{Enjoying the baseball game from the third-base
%  seats. Ichiro Suzuki preparing to bat.}
%  \label{fig:teaser}
%\end{teaserfigure}

%%
%% This command processes the author and affiliation and title
%% information and builds the first part of the formatted document.
\maketitle

%%%%%%%%%%%%%%%%%%%%%%%%%%%%%%%%%%%%%%%%%%%%

\section{Introduction}
\label{sec:intro}

One of the key reasons for the success of metaheuristics is their being general-purposeness. Indeed, Evolutionary Algorithms (EAs), Swarm Intelligence (SI) algorithms and alike can be applied, more or less straightforwardly, to a broad range of optimization problems. On the other hand, it is well-established that different algorithms can produce different results on a given problem, and in fact it is impossible to identify an algorithm that works better than any other algorithm on all possible problems \cite{wolpert1997no}.

Moreover, the performance of metaheuristics typically depends on their hyper-parameters. However, optimal parameters are usually problem-dependent, and finding those parameters before performing an optimization process through trial-and-error or other empirical approaches is usually tedious, and obviously suboptimal. One possible alternative is given by hyperheuristics \cite{burke2013hyper,drake2020recent,sanchez2020systematic}, i.e., algorithms that can either select the best metaheuristic for a given problem \cite{nareyek2003choosing,chakhlevitch2008hyperheuristics,li2017learning}, or simply optimize the parameters of a given metaheuristic. Several tools, e.g. \texttt{irace} \cite{lopez2016irace}, exist for this purpose.

% \cite{burke2003hyper} {Hyper-heuristics: An emerging direction in modern search technology}
% \cite{ross2005hyper} {Hyper-heuristics}
% \cite{ozcan2008comprehensive} {A comprehensive analysis of hyper-heuristics}

Another possibility is to endow the metaheuristic with a parameter \emph{adaptation} strategy, i.e., a set or rules that change the parameters dynamically during the optimization process. Several handcrafted, successful policies have been proposed over the years to address parameter adaptation \cite{cotta2008adaptive}. However, finding an optimal adaptation policy is, in turn, challenging as different policies may perform differently on different problems or during different stages of an optimization process. Moreover, exploring manually the space of such policies is infeasible. On the other hand, it is possible to cast the search for an adaptation policy as a reinforcement learning (RL) problem \cite{sutton2018reinforcement}, where the agent observes the state of the optimization process and decides how to change the parameters accordingly. However, only few attempts have been done so far in this direction. This is mostly due to the fact that the observation space of an optimization process can be quite large, and finding relevant state metrics (i.e., inputs to the policy) and rewards can be difficult.

Here, we aim to make steps in this direction by introducing a general-purpose framework for performing parameter adaptation in continuous-domain metaheuristics based on state-of-the-art RL. One reason for building such a framework is to relieve algorithm designers and practitioners from the need for building handcrafted adaptation strategies. Moreover, using such framework would allow to use pretrained strategies and apply them to new optimization problems.

In the experimentation, we focus on two well-known continuous optimization algorithms (assuming, without loss of generalization, minimization of the objective function/fitness), namely the Covariance Matrix Adaptation Evolution Strategies (CMA-ES) \cite{hansen2001completely} and Differential Evolution (DE) \cite{storn1997differential}, for which well-known successful handmade adaptation policies exist. In the case of CMA-ES, we train an adaptation policy for the step-size $\sigma$. In the case of DE, we instead adapt the scale factor $F$ and the crossover rate $CR$. We train these policies on a set of 46 benchmark functions at different dimensionalities, with various state metrics, in two settings: one policy per function, and one global policy for all functions. Compared, respectively, to the Cumulative Step-size Adaptation (CSA) policy \cite{chotard2012cumulative} and to two well-known adaptive DE variants (iDE 
\cite{elsayed2011differential} and jDE \cite{brest2006self}), our policies are able to produce competitive results, especially in the case of DE.

The rest of the papers is structured as follows. In the next section, we briefly present the related works. In Section \ref{sec:methods}, we describe our methods. In Section \ref{sec:results}, we then present our results. Finally, we draw the conclusions in Section \ref{sec:conclusions}.

%%%%%%%%%%%%%%%%%%%%%%%%%%%%%%%%%%%%%%%%%%%%

\section{Background}
\label{sec:background}

In the context of DE, several works have shown the effect of using an adaption strategy to choose $F$ and $CR$. These parameters are, in fact, known to affect both diversity and optimization results \cite{yaman2019comparison}. For instance, some authors proposed using pools of different parameters and mutation/crossover strategies, either as discrete sets of fixed values \cite{iacca2014differential}, or as continuous ranges \cite{iacca2015continuous}. Others proposed using multiple mutation strategies \cite{yaman2018multi}, where each strategy is represented as an agent whose measured performance is used to promote its activation within an ensemble of strategies. Recently, the authors of \cite{blank2022} introduced a polynomial mutation for DE with different approaches for controlling its parameter. The authors of \cite{ghosh2022} proposed instead an improvement on SHADE \cite{tanabe2013success}, which uses proximity-based local information to control the parameter settings.

Rather than engineering the parameter adaptation strategy, some studies have tried to learn metaheuristics with RL. Some of these works are based on Q-learning: Li et al. \cite{li2019differential} considered each individual as an agent that learns the optimal strategy for solving a multi-objective problem with DE; in a similar way, Hu et al. \cite{hu2021reinforcement} used a Q-table for each individual to choose how much to increase/decrease the $F$ parameter during a DE run to solve circuit design problems; Sallam et al. \cite{sallam2020evolutionary} proposed an algorithm that evolves two populations: one with CMA-ES, and one with Q-table, in order to choose between different DE operators and enhance the EA with a local search.

Other approaches are based on deep RL: Sharma et al. \cite{sharma2019deep} proposed a method that uses deep RL that produces an adaptive DE strategy based on the observation of several state metrics; Sun et al. \cite{sun2021learning} trained a Long-Short Term Memory (LSTM) with policy gradient to control the $F$ and $CR$ parameters in DE; Shala et al. \cite{shala2020learning} trained a neural network with Guided Policy Search (GPS) \cite{levine2013guided} to control the step-size of CMA-ES by also sampling trajectories created by Cumulative Step-size Adaptation (CSA) \cite{chotard2012cumulative}; Lacerda et al. \cite{lacerda2021out} used distributed RL to train several metaheuristics with Twin Delayed Deep Deterministic Policy Gradients \cite{fujimoto2018addressing}.

%It is worth to mention also some application of EAs applied to deep learning: \cite{real2020automl} makes use of an evolutionary search to evolve a ML algorithm; \cite{sebag2017stochastic} uses only the CSA strategy to update layer-wise learning rates in a neural network.

%%%%%%%%%%%%%%%%%%%%%%%%%%%%%%%%%%%%%%%%%%%%

\section{Methods}
\label{sec:methods}
The proposed framework uses deep RL to learn parameter adaptation strategies for EAs, i.e., to learn a policy that is able to set the parameters of an EA at each generation of the optimization process. In that, our framework is similar to the approach presented in \cite{shala2020learning}. However, differently from \cite{shala2020learning} we do not use GPS as RL algorithm and, most importantly, we do not partially sample the parameter adaptation trajectory from an existing adaptation strategy (in \cite{shala2020learning}, CSA), but rather we build the adaptation trajectory from scratch, i.e., entirely based on the trained policy. Another important aspect is that our framework can be configured with different EAs and RL algorithms, and can be easily extended in terms of state metrics, actions and rewards.

Next, we briefly describe the two EAs considered in our experimentation (Section \ref{sec:eas}), the RL setting (Section \ref{sec:rlmodel}), the evaluation procedure (Section \ref{sec:evaluation}) and the computational setup (Section \ref{sec:setup}).

\subsection{Evolutionary algorithms}
\label{sec:eas}
We tested the framework using CMA-ES and DE since these are two well-known EAs for which several studies exist on parameter adaptation. In our comparisons, we considered two well-established adaptation strategies taken from the literature: for CMA-ES, Cumulative Step-size Adaptation (CSA) \cite{chotard2012cumulative}; for DE, iDE \cite{elsayed2011differential} and jDE \cite{brest2006self}. More details on these adaptation strategies will follow.

\subsubsection{Covariance Matrix Adaptation Evolution Strategies}
CMA-ES \cite{hansen2001completely} conducts the search by sampling adaptive mutations from a multivariate normal distribution ($x_i\sim \boldsymbol{m} + \sigma \times \mathcal{N}(0,\boldsymbol{C})$). At each generation, the mean $\boldsymbol{m}$ is updated based on a weighted average over the population, while the covariance matrix $\boldsymbol{C}$ is updated by applying a process similar to that of Principle Component Analysis. The remaining parameter, $\sigma$ is the step size, which in turn is adapted during the process. Usually, $\sigma$ is self-adapted using CSA \cite{chotard2012cumulative}. In our case, the policy is learned and computed based on an observation of the current state of the search.

\subsubsection{Differential Evolution}
DE \cite{storn1997differential} is a very simple yet efficient EA. Starting from an initial random population, at each generation the algorithm applies on each parent solution a differential mutation operator, to obtain a \emph{mutant}, which is then crossed over with the  parent. While there are different mutation and crossover strategies for DE, in this study we consider only the ``best/1/bin'' strategy. According to this strategy, the mutant is computed as $\boldsymbol{x}^{k+1} = \boldsymbol{x}^k_{best} + F \times (\boldsymbol{a}-\boldsymbol{b})$; where $\boldsymbol{x}^k_{best}$ is the best individual at the $k$-th generation, $\boldsymbol{a}$ and $\boldsymbol{b}$ are two mutually exclusive randomly selected individuals in the current population, and $F$ is the scale factor. The binary crossover, on the other hand, swaps the genes of parent and mutant with probability given by the crossover rate $CR$.

Without adaptation, $F$ and $CR$ are fixed. In our case, we make the policy learn how to adapt them by using two different approaches: directly updating $F$ and $CR$ with the policy, or sampling $F$ and $CR$ from a uniform/normal distribution parametrized by the policy.

\subsection{Reinforcement learning setting}
\label{sec:rlmodel}
As for the RL setting, we chose the same model used in \cite{shala2020learning}: 2 fully connected hidden layers of 50 neurons each (thus with $50 \times 50$ connections) with ReLU activation function. The size if the input layer depends on the observation space, while the size of the output layer depends on the action space. In the following, we describe the other details of the learning setting.

\subsubsection{Proximal Policy Optimization}
\label{ppo}
We chose Proximal Policy Optimization (PPO) \cite{ppo} to optimize the policy due to its good performances in general-purpose RL tasks. Here, for brevity we do not go into details of the algorithm (for which we refer to \cite{ppo}), but in short the algorithm works as shown in Algorithm \ref{alg:cap}.

% \begin{equation}
%     L_t(\theta)= \hat{\mathbb{E}}_t\Big[L^{CLIP}_t(\theta)-c_1L^{VF}_t+c_2S[\pi_\theta](s_t)\Big]
% \end{equation}

In our setup, $N=1$, $K=200$, and the other parameters are set as per their defaults value used in the \texttt{ray-rllib} library\footnote{See \url{https://docs.ray.io/en/latest/rllib/rllib-algorithms.html\#ppo}}. $\theta$ are the parameters of the policy (in our case, the weights of the neural networks), $L$ is the loss function (see Eq. 9 from \cite{ppo}) and $\hat{A}_t$ is the advantage estimate at iteration $t$ (see Eq. 11 from \cite{ppo}).

\subsubsection{Observation spaces}
\label{obSpaces}
We experimented with different observation spaces, each one defined as a set of \textit{state metrics}. A state metric computes the state (or observation) of the model based on various combinations of fitness values, genotypes, and other parameters of the EA. More specifically, we used the following state metrics:

\begin{itemize}[leftmargin=*]
    \item \textit{Inter-generational $\Delta f$:} For the last $g$ generations, we take the best fitness in the population at each generation and compute the normalized difference with the best fitness at the previous generation:
    \begin{gather}
        \Delta f^{inter}_g = \Big[\Delta f^{inter}_k, \Delta f^{inter}_{k-1}, \dots, \Delta f^{inter}_{k-(g-1)}\Big] \\
        \text{where} \quad \Delta f^{inter}_k = \frac{f^*_k - f^*_{k-1}}{|f^*_k - f^*_{k-1}| + |f^*_{k-1}| + 10^{-5}}
        \label{interdeltaf}
    \end{gather}
    where $f^*_k$ is the best fitness value in the population at the $k$-th generation. In this way, $\Delta f^{inter}_k \in (-1, 1)$ and it is proportional to the best fitness from the previous generation, saturating to $\pm 1$ for $f^{*}_k \rightarrow \pm \infty$. The constant $10^{-5}$ is needed to avoid divisions by zero. The normalization of $\Delta f$ is fundamental to have stable training.
    %using different RL algorithms or object functions.
    \item \textit{Intra-generational $\Delta f$:} For the last $g$ generations, we take the normalized difference between the maximum and minimum fitness of the current population at each generation:
    \begin{equation}
        \Delta f^{intra}_k = \frac{|f^{max}_k - f^{min}_k|}{|f^{max}_k - f^{min}_k|+|f^*_k|+10^{-5}}.
    \end{equation}
    \item \textit{Inter-generational $\Delta \boldsymbol{X}$:} Similarly to the inter-generational $\Delta f$, the normalized difference between the best genotype in two consecutive generations are taken for the last $g$ generations. In this case, to maintain linearity, the normalization is done using the bounds of the search space:
    \begin{equation}
        \Delta \boldsymbol{X}^{inter}_k = \frac{\boldsymbol{X}^*_k - \boldsymbol{X}^*_{k-1}}{\Delta \boldsymbol{X}^{bounds}}
    \end{equation}
    where $\boldsymbol{X}^*_k$ is the genotype associated to the best fitness at generation $k$ and $\Delta \boldsymbol{X}^{bounds} = \Big[\Delta \boldsymbol{X}^{bounds}_1,\dots,\Delta \boldsymbol{X}^{bounds}_d\Big]$ is the vector containing, for each variable, the bounds of the search space, being $d$ the problem size.
    Since the size of this observation would depend on the problem size, the policy would work only with problems of that fixed size. To solve this problem, we use as observation the minimum and maximum values of $\Delta \boldsymbol{X}^{inter}_k$:
    \begin{equation}
        \Delta \boldsymbol{X}^{inter}_{k_{min,max}}=\Big[\min(\Delta \boldsymbol{X}^{inter}_k),\max(\Delta \boldsymbol{X}^{inter}_k) \Big].
    \end{equation}
    The intra-generational $\Delta \boldsymbol{X}$ is then defined as a history of the above defined metric at the last $g$ generations: 
    \begin{equation}
        \Delta \boldsymbol{X}^{inter} = \Big[ \Delta \boldsymbol{X}^{inter}_{k_{min,max}},\Delta \boldsymbol{X}^{inter}_{k-1_{min,max}},\dots,\Delta \boldsymbol{X}^{inter}_{k-(g-1)_{min,max}} \Big].
    \end{equation}
    \item \textit{Intra-generational $\Delta \boldsymbol{X}$:} Given $\boldsymbol{X}^k_{i_j}$ as the $j$-th dimension of the $i$-th individual of the population at the $k$-th generation, the intra-generational $\Delta \boldsymbol{X}$ at the $k$-th generation is defined as:
    \begin{gather}
        \Delta \boldsymbol{X}^{intra}_k = \Big[ \boldsymbol{X}^{intra}_{k_1},\dots,\boldsymbol{X}^{intra}_{k_d}\Big]\\
        \text{where} \quad \Delta \boldsymbol{X}^{intra}_{k_j} = \frac{\left|\boldsymbol{X}^k_{max_j} - \boldsymbol{X}^k_{min_j}\right|}{\Delta \boldsymbol{X}^{bounds}_j}.
    \end{gather}
    Also in this case, we use as observation the minimum and maximum values of $\Delta \boldsymbol{X}^{intra}_k$:
    \begin{equation}
        \Delta \boldsymbol{X}^{intra}_{k_{min,max}}=\Big[\min(\Delta \boldsymbol{X}^{intra}_k),\max(\Delta \boldsymbol{X}^{intra}_k) \Big].
    \end{equation}
    The intra-generational $\Delta \boldsymbol{X}$ is then defined as a history of the above defined metric at the last $g$ generations: 
    \begin{equation}
        \Delta \boldsymbol{X}^{intra} = \Big[ \Delta \boldsymbol{X}^{intra}_{k_{min,max}},\Delta \boldsymbol{X}^{intra}_{k-1_{min,max}},\dots,\Delta \boldsymbol{X}^{intra}_{k-(g-1)_{min,max}} \Big].
    \end{equation}
\end{itemize}
In all the experiments, we always include in the observation space also the \textit{previous model output}, i.e., the parameters given by the model in the previous generation.

\begin{algorithm}[ht!]
    \caption{High-level description of PPO.}\label{alg:cap}
    \For{iteration = $1, 2, \dots I$}{
        \For{actor = $1, 2, \dots N$}{
            Run policy $\pi_{\theta_{old}}$ in environment for $T$ timesteps \\
            Compute advantage estimates $\hat{A}_1,\dots,\hat{A}_T$
        }
        Optimize loss $L$ w.r.t. $\theta$, with $K$ epochs and minibatch size $M \leq N\times T$ \\
        $\theta_{old} \leftarrow \theta$
    }
\end{algorithm}

\subsubsection{Action spaces}
\label{actions}
The action space of the policy depends on both the specific EA and the approach used to parametrize it. In our model, the action is taken at every generation, using the observation from the previous one. In our experiments, we considered the following action spaces:
\begin{itemize}[leftmargin=*]
    \item \textit{CMA-ES (Step-size)}: $\sigma \in [10^{-10},3]$.
    % \item \textit{$F$-only in DE}: $F \in [0,2]$
    % \item \textit{$CR$-only in DE}: $CR \in [0,1]$
    \item \textit{DE (Both $F$ and $CR$)}: $F \in [0,2], CR \in [0,1]$.
    \item \textit{DE (Normal distribution)}: $F$ and $CR$ are sampled using two normal distributions parametrized with mean and standard deviation determined by the learned policy, i.e., respectively, $\mathcal{N}(\mu_F,\sigma_F)$ and $\mathcal{N}(\mu_CR,\sigma_CR)$. Thus, the action space is: $\Big\{\mu_F \in [0,2]$, $\sigma_F \in [0,1]$, $\mu_{CR} \in [0,1]$, $\sigma_{CR} \in [0,1]\Big\}$.
    \item \textit{DE (uniform distribution)}: $F$ and $CR$ are sampled using two uniform distributions parametrized with lower and upper bound determined by the learned policy, i.e., respectively, $\mathcal{U}(F_{min}, F_{max})$ and $\mathcal{U}(CR_{min},CR_{max})$. Thus, the action space is: $\Big\{F_{min} \in [0,2]$, $F_{max} \in [0,2], CR_{min} \in [0,1], CR_{max} \in [0,1]\Big\}$.
\end{itemize}

\subsubsection{Reward}
\label{reward}
The reward is a scalar representing how good or bad was the performance of the policy during the training episodes (in our case, an episode is a full evolutionary run). It is computed every generation using the \textit{Inter-generational $\Delta f$} without history, see Eq. \eqref{interdeltaf}. The use of this reward brings some advantages: it reflects the progress of the optimization process, maintaining the independence with different scales of the objective functions, and it yields better numerical stability during the training process. All the experiments have been done using this reward function (except the one presented in Section \ref{PPOvsGPS}).

\subsubsection{Training procedure}
\label{training}
We consider two policy configurations: single-function policy, and multi-function policy. In the first configuration, the model is trained separately on each function: in this way, the policy specializes for each single optimization problem. If the ideal policy should work well on as many functions as possible, a single-function policy can be useful to get an idea about the top performance that the policy can get for each function. Or, single-function policies could used for similar functions. In the second configuration, the model is instead trained using the evolutionary runs on multiple functions. Quite surprisingly, with this procedure, we could obtain policies that are able to work better than the adaptive approaches from the literature.

In all our experiments, we trained the models for $5000$ function evolutionary runs (i.e., episodes), each consisting of $500$ function evaluations, hence meaning $2.5 \times 10^6$ function evaluations per each policy training. Then, the trained policy is tested using the procedure defined in Section \ref{testingproc}.

\subsection{Evaluation}\label{sec:evaluation}

\subsubsection{Benchmark functions}
The experiments have been done with 46 benchmark functions taken from the BBOB benchmark \cite{bbob}. For each function, we used the default instance, i.e., without random shift in the domain or codomain (as done in \cite{shala2020learning}). Future investigations will extend the analysis to instances with shift.

The 46 functions are selected as follows. The first 10 functions are: BentCigar, Discus, Ellipsoid, Katsuura, Rastrigin, Rosenbrock, Schaffers, Schwefel, Sphere, Weierstrass, all in 10 dimensions. The remaining 36 functions are the same 12 functions, namely: AttractiveSector, BuecheRastrigin, CompositeGR, DifferentPowers, LinearSlope, SharpRidge, StepEllipsoidal, RosenbrockRotated, SchaffersIllConditioned, LunacekBiR, GG101me, and GG21hi, each one in 5, 10 and 20 dimensions. 

\subsubsection{Compared methods}
We compared the learned policies with the following adaptive methods from the literature:
\begin{enumerate}[leftmargin=*]
    \item Cumulative Step-size adaptation \cite{chotard2012cumulative}:
    CSA is considered the default step-size control method of CMA-ES. To compute $\sigma_{t+1}$, a cumulative path is defined as: $\boldsymbol{p}_{t+1}=(1-c)\boldsymbol{p}_t+\sqrt{c(2-c)}\boldsymbol{\xi}^*_t$, where $c\in[0,1]$ ($1/c$ represents the lifespan of the information contained in $\boldsymbol{p}_t$) and $\boldsymbol{\xi}^*_t$ is the best children at the $t$-th generation. The step-size is defined as: $\sigma_{t+1}= \sigma_t exp(\frac{c}{d_\sigma}(\frac{||\boldsymbol{p}_{t+1}||}{E(||\mathcal{N}(0,I_n)||)}-1))$, where $d_\sigma$ is the damping parameter that determines how much the step size can change (usually, $d_\sigma=1$).
    \item iDE \cite{elsayed2011differential}:
    The iDE adaptive method maintains a different $F$ and $CR$ for each individual and updates them with a different rule that depends on the mutation/crossover strategy used. Since in our DE experiments we use the best/1/bin strategy, the considered iDE update rules are:
    \begin{gather}
        F = F_{best} + \mathcal{N}(0,0.5)\cdot(F_{r_1}-F_{r_2}) \\
        CR = CR_{best} + \mathcal{N}(0,0.5)\cdot(CR_{r_1}-CR_{r_2})
    \end{gather}
    where $F_{best}$ and $CR_{best}$ are the $F$ and $CR$ values corresponding to the best individual and $F_{r_i}$ (or $CR_{r_i}$) is a random $F$ (or $CR$) sampled from the best $F$ (or $CR$) values until the current generation ($i$ is needed to selected mutually exclusive values for each individual).
    \item jDE \cite{brest2006self}:
    jDE is a simple but effective adaptive DE variant. With probability $p=0.1$ the method samples $F$ from $\mathcal{U}(0.1,1)$. Otherwise, it uses the best $F$ until the current generation.
\end{enumerate}

\subsubsection{Evaluation metrics}
\label{evalmetrics}
In order to compare the different setups of algorithms and models, we consider two metrics (similar to \cite{shala2020learning}):
\begin{itemize}[leftmargin=*]
    \item \textit{Area Under the Curve (AUC)}:
    During each evolutionary run, the minimum fitness of the population at each generation is stored. The result is a monotonic non-increasing discrete function (assuming elitism). The area under this curve is then calculated using the composite trapezoidal rule. This metric is a good indication of how fast the optimization process is.
    \item \textit{Best of Run}:
    The best fitness found during the entire optimization process.
\end{itemize}
As we assume minimization of the objective function, for both metrics it holds that the lower their values, the better is the performance of a policy.

\subsubsection{Testing procedure}
\label{testingproc}
Given a RL based trained policy $\pi_A$ and an adaptive policy $\pi_B$ taken from the literature (e.g., CSA) that take actions on the corresponding EA (e.g., CMA-ES), the two policies are tested in the following way:
\begin{enumerate}[leftmargin=*]
    \item We take the policy $\pi_A$ and execute 50 runs, each one for 50 generations, with a population of 10 individuals. Thus, every run has 500 function evaluations.
    \item We do the same for policy $\pi_B$.
    \item For each run of both policies, we compute the two metrics (AUC and Best of Run).
    \item For both metrics, we calculate the probability that $\pi_A$ performs better than $\pi_B$ as:
    \begin{equation}
        \label{eq:probcomp}
        p(\pi_A < \pi_B)=\frac{1}{n^2}\sum_i^n\sum_j^n\mathbb{1}_{\pi_{A_i}<\pi_{B_j}}
    \end{equation}
    where $\mathbb{1}_{\pi_{A_i}<\pi_{B_j}}$ is 1 if the metric of $\pi_A$ on the $i$-th run is less than the metric of $\pi_B$ on the $j$-th run, otherwise it is 0.
\end{enumerate}

\subsection{Computational setup}\label{sec:setup}
We ran our experiments on an Azure Virtual Machine with an 8 core 64-bit CPU (we noted that the CPU model would change over different sessions, but usually the machine used an Intel Xeon with $>$2GHz and $>$30MB cache) and 16GB RAM, running Ubuntu 20.04. A training process of $5000$ episodes takes $\sim 1-1.5$ hours.

Our code is implemented in python (v3.8) using \texttt{ray-rllib} (v1.7), \texttt{gym} (v0.19) and \texttt{numpy} (v1.19). We took the CSA implementation from the \texttt{cma} (v3.1.1) library, as well as the BBOB benchmark functions \cite{bbob}. The implementation of DE was done by slightly modifying the \texttt{scipy}'s implementation in order to make it compatible with $F$ and $CR$ at the individual level. The implementation of iDE and jDE has been realized by porting it from the C++ implementation available in the \texttt{pagmo} (v2.18.0) library (that is based on the algorithm descriptions presented in \cite{elsayed2011differential} and \cite{brest2006self}).

%%%%%%%%%%%%%%%%%%%%%%%%%%%%%%%%%%%%%%%%%%%%

\section{Results}
\label{sec:results}

We now present the results, separating the experiments with CMA-ES (Section \ref{results:cmaes}) from those with DE (Section \ref{results:de}).

\subsection{CMA-ES experiments}\label{results:cmaes}

\subsubsection{Comparison between PPO and GPS}
\label{PPOvsGPS}
The first experiment was done with CMA-ES and single-function training, trying to configure the model as similarly as possible to \cite{shala2020learning}, in order to get a first comparative analysis. However, a direct comparison with the results reported in \cite{shala2020learning} was not possible. In fact, the authors of \cite{shala2020learning} used GPS as training algorithm, that is not implemented in the \texttt{ray-rllib} library. To avoid replicability issues, we then decided to train our model using the available PPO implementation from \texttt{ray-rllib}. Furthermore, we did not use the sampling rate technique implemented in \cite{shala2020learning}, i.e., in our case the trajectories of the step-size are taken entirely from the trained policy.

The rest of the setup is the same used in \cite{shala2020learning}. As mentioned earlier, we used 2 fully connected hidden layers of 50 neurons each with ReLU. The observation space is: differences between successive fitnesses from 40 previous generations (not normalized), the step-size history from 40 previous generations, the current cumulative path length (Equation 2 from \cite{shala2020learning}). The reward is the negative of the fitness (not normalized). The action space is $\sigma \in [0.05,3]$. Please note that these state metrics and reward are different from the ones described in Section \ref{sec:rlmodel}, and have been used only in this preliminary experiment for comparison with the results from \cite{shala2020learning}.  

We performed this experiment only with the first 10 functions of the considered benchmark. The result of this experiment was quite poor: the single-function trained policy obtained better testing results than CSA ($p(\pi<CSA)$ with both AUC and Best of Run metrics) only on 2 functions. We found that the main reason for this scarce performance is the noisy reward. In fact we observed that, depending on the function, the scale of the fitness differs across multiple runs, and PPO is sensible to the reward scale. 
% (especially the clip parameter of the value function)
This seems to explain why the authors of \cite{shala2020learning} chose GPS, which is robust to different reward scales.

Also, with this setup we encountered numerical instability problems: with BentCigar, Rosenbrock and Schaffers we have not been able to train the policy because at a certain point of the training process the weights of the model became NaN. This is very likely caused by the noisy reward, which makes some gradient or loss function value go to infinity. Indeed, this problem was almost totally fixed using a normalized reward.

% \begin{table}[h!]
%     \centering
%     \begin{tabular}{ |p{1.5cm}||p{2.5cm}|p{2.5cm}|  }
%         \hline
%         Function & $p(PPO < CSA)$ with AUC metric & $p(PPO < CSA)$ with Best of Run metric \\
%         \hline
%         bentcigar   & \textbf{0.58} & 0.17 \\
%         % discus      & --- & --- \\
%         ellipsoid   & 0.35 & 0.22 \\
%         katsuura    & 0.50 & \textbf{0.54} \\
%         rastrigin   & \textbf{1} & \textbf{0.95} \\
%         % rosenbrock  & --- & --- \\
%         % schaffers   & --- & --- \\
%         schwefel    & \textbf{0.94} & 0.27 \\
%         sphere      & 0.0 & 0.0 \\
%         weierstrass & 0.08 & 0.01 \\
%         \hline
%         Total       & 43\% (3/7) & 29\% (2/7) \\
%         \hline
%     \end{tabular}
%     \caption{results with the configuration as similar as possible to \cite{shala2020learning}}
%     \label{table:cmapaper}
% \end{table}

\subsubsection{Normalizing the reward}
We tried to improve the previous setup by normalizing the reward and using a minimal observation space. The reward in this case is the one explained in Section \ref{reward}. The observation space is the inter-generational $\Delta f$ with $g=40$ and the step-size of the previous generation. Testing the policy on all the 46 functions, it did better than CSA on 30.4\% (14/46) of the functions. Moreover, we did not have training stability issues. Overall, we found that CSA is a very good step-size adaptation strategy and it is difficult to do better by means of RL.

\subsection{DE experiments}\label{results:de}
A more intensive experimentation has been conducted with DE. We started with single-function training policies, training one model per function. Then we experimented with multi-function training, applying small changes in the model in order to get close to the single-function results.

\subsubsection{Single-function policy}
In Figure \ref{fig:DEsingle} we report the results of the single-function training policies using three action spaces to parametrize DE, and compare them with iDE and jDE. For brevity, we report only the results of the Best of Run metric. Green (red) cells indicate that the trained policy $\pi$ works better (worse) than the corresponding adaptive DE variant (thus, either iDE or jDE), with $\pi$ trained separately and tested on each function. Darker green (red) indicates higher (lower) probabilities. Black cells indicate that the policy could not be trained due to numerical instability issues: in fact DE, due to its random nature, is likely to produce different fitness trajectories across evolutionary runs. This causes a noisy reward that can lead to numerical instabilities during the training process.

To solve this problem, it is necessary to design a custom loss function for the training algorithm. However, this would mean to use a variation of PPO, which falls outside the scope of this work where we are limiting ourselves to using the original PPO. A simple workaround was to run the training process multiple times: in most cases, one or two attempts were enough to train the policy without encounter this instability problem. Moreover, we observed that using the hyperbolic tangent as activation function (instead of ReLu) can help reduce the probability to encounter instabilities. However, we did not perform a deeper analysis on this.

The leftmost side of Figure \ref{fig:DEsingle} shows the percentage of the functions where the learned policy did better than iDE/jDE over the total number of functions: $ratio = \frac{||p(\pi_i<iDE/jDE)>0.5||}{n_{functions}}$. It can be seen that the uniform distribution strategy gives the best results overall. However, there are a few functions where the adaptive strategies provided by iDE and jDE always do better.

\subsubsection{Multi-function policy}
After seeing the results of the normal and uniform distribution approaches in the single-function setting, we experimented with multi-function training using one policy trained for $5000$ episodes on all the $46$ functions, meaning $\sim 108$ evolutionary runs per function. We trained and compared 9 versions of the model with different observation spaces combining the state metrics defined in Section \ref{obSpaces}. 

\begin{figure*}[ht!]
    \includegraphics[width=\textwidth]{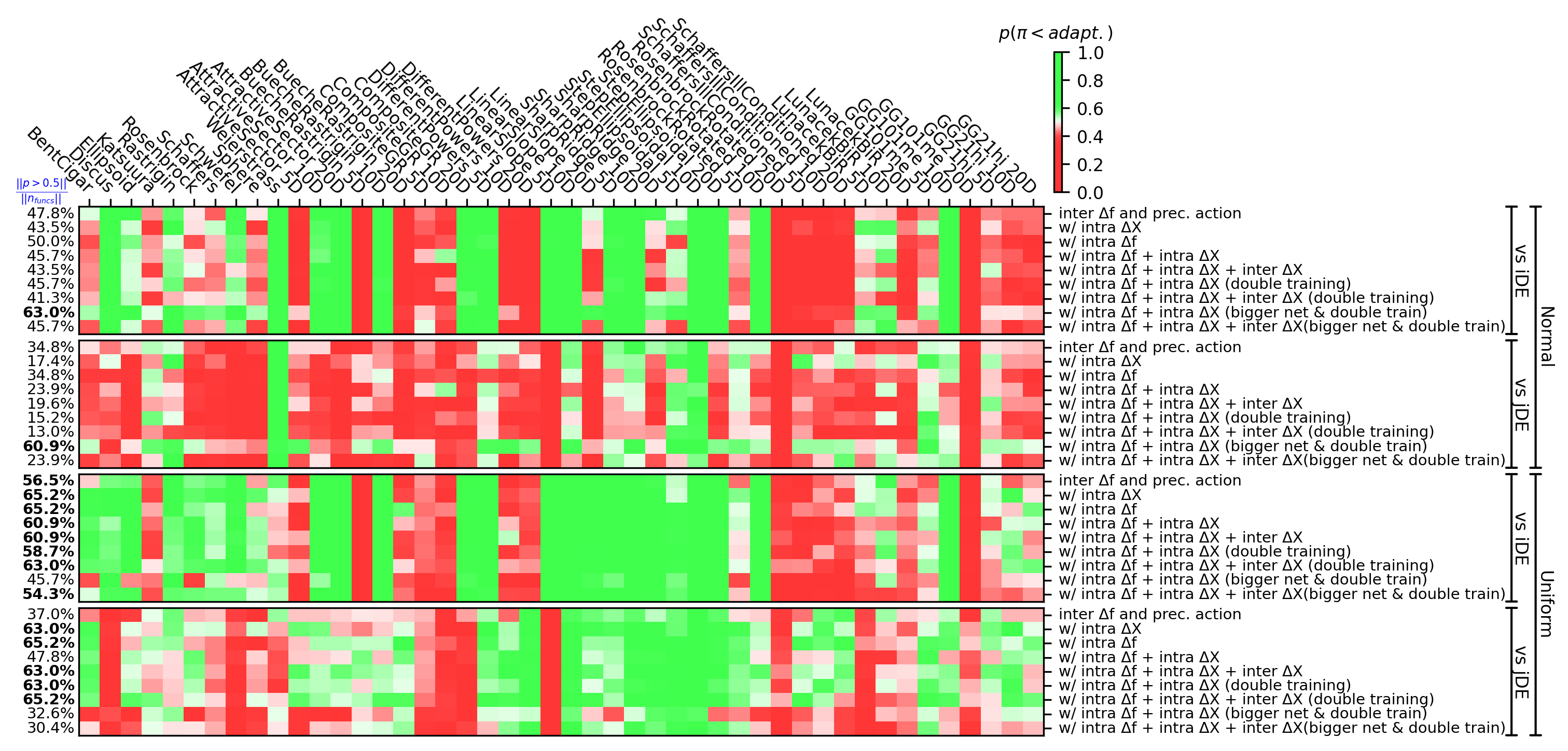}
    \vspace{-0.8cm}
    \caption{Multi-function training policies with different observation spaces and training times, compared with iDE (first and third row) and jDE (second and fourth row). The color is based on $p(\pi<adapt.)$, see Eq. (\ref{eq:probcomp}), calculated on the ``Best of Run'' metric, where $\pi$ is the policy trained on all functions and tested on each function, and ``adapt.'' is either iDE or jDE. In the first two rows, the normal distribution approach is used; in the last two rows, the uniform distribution approach is used. Green (red) cells indicate that the trained policy $\pi$ works better (worse) than the corresponding adaptive DE variant. Darker green (red) indicates higher (lower) probabilities. Percentages on the left side of the rows are calculated as $\frac{\text{no. green cells}}{\text{no. green cells+no. red cells}}$ on the same row.}
    \label{fig:DEmulti}
\end{figure*}

The results of this experiment are shown in Figure \ref{fig:DEmulti}. All the policies have at least the inter-generational $\Delta f$ and the values of the precedent action as observation. The entries on the rightmost side of Figure \ref{fig:DEmulti} (starting with ``w/'') denote what is included in the observation space. Moreover, we also tried to double the number of training episodes (``double training'' labels) and increase the size of the model to $100 \times 50 \times 10$ (``bigger net'' label).

\begin{figure*}[ht!]
    \includegraphics[width=\textwidth]{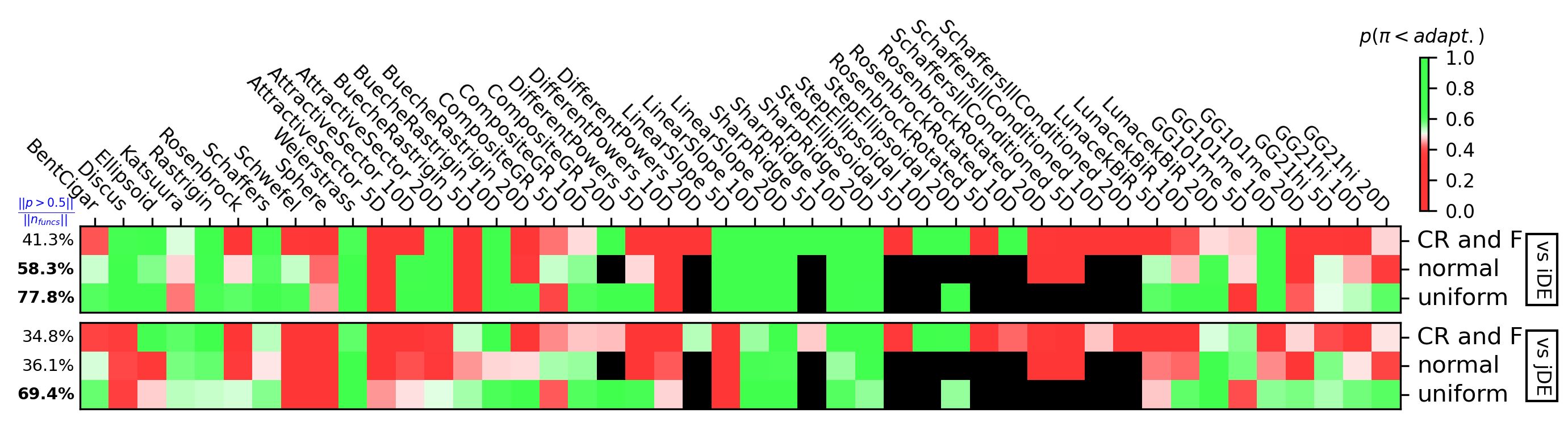}
    \vspace{-0.8cm}
    \caption{Single-function training policies compared with iDE (top) and jDE (bottom). The color is based on $p(\pi<adapt.)$, see Eq. (\ref{eq:probcomp}), calculated on the ``Best of Run'' metric, where $\pi$ is the policy trained separately and tested on each function, and ``adapt.'' is either iDE or jDE. Black cells indicate that the model has not been trained due to numerical instability. Green (red) cells indicate that the trained policy $\pi$ works better (worse) than the corresponding adaptive DE variant. Darker green (red) indicates higher (lower) probabilities. Percentages on the left side of the rows are calculated as $\frac{\text{no. green cells}}{\text{no. green cells+no. red cells}}$ on the same row.}
    \label{fig:DEsingle}
\end{figure*}

One of the main results that can be noted from Figure \ref{fig:DEmulti} is the different performance between the normal and the uniform distribution approach. The latter is visibly superior with respect to the former, and it gets very close to the single-function training performance shown in Figure \ref{fig:DEsingle}, by only adding the intra-generational $\Delta f$ to the observation space. The normal distribution approach overcomes iDE and jDE only by both adding intra-generational $\Delta f$ and intra-generational $\Delta \boldsymbol{X}$ to the observation space and increasing the model and its training time. This suggests the fact that this approach could work but it is more difficult to train. Another consideration may be that, in order to get a better balance between exploration and exploitation, $F$ and $CR$ must be highly variant, especially at the end of the evolution.
%So by increasing the standard deviation ranges in the action space of the normal distribution approach, it can allow the policy to get better performances in the optimization process.

\begin{figure}[ht!]
    \includegraphics[scale=0.6]{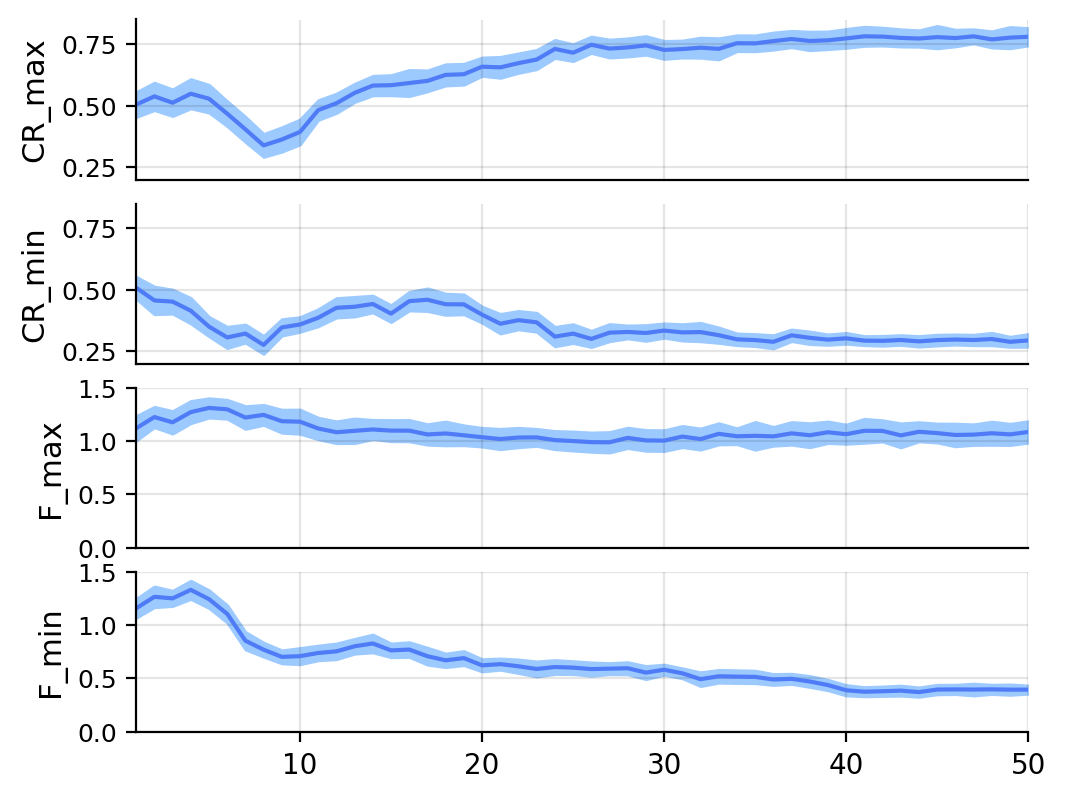}
    \caption{Action trajectories produced by the ``w/ intra $\Delta f$ uniform distribution'' policy (mean $\pm$ std. dev. across $46$ benchmark functions). Given a tensor of shape $(46,50,4,50)$, storing $4$ actions of the policy during $50$ generations of $50$ evolutionary runs for each of the $46$ functions, we compute first the mean across runs, then the mean and std. dev. across functions.}
    \label{fig:actions}
\end{figure}

Another important observation can be made looking at Figure \ref{fig:actions}. Because the trajectories are very similar across the functions (note that there is a small standard deviation in the actions), it is clear that the policy is not able to differentiate trough different functions, but rather it learns to map actions to the number of the current generation. Figure \ref{fig:actions} shows only the trajectories of one policy, but the same pattern is present on all the multi-function training policies. This problem is an important limitation for the policy because being able to make different choices depending on the function is crucial if we want to have true adaptation. One possible cause of this problem may be the small capacity of the model (in terms of number of layers/neurons) or the number of episodes. However, increasing both (at least to the values that we tested) did not bring any improvement. A possible solution may be to add a loss in an intermediate layer (like GoogLeNet \cite{szegedy2015going} does) to classify the function in some manner (e.g., unimodal or multimodal). However, this would increases the computational cost.

Figure \ref{fig:actions} also shows that the model has learned a general policy that works well for the majority of the functions and that this is in line with the common strategy of: exploration during the first phase, and exploitation during the last phase. In fact, during the evolution, $CR$, which determines the effect of the crossover, is initially small and within a small range ($CR_{max}$ and $CR_{min}$ are $<0.5$ and similar) while at the end it increases its variance ($CR_{max}=0.75$ and $CR_{min}=0.25$). Instead, $F$, which determines the effects of mutation, is initially high with low variance (both $F_{max}$ and $F_{min}$ are $\sim 1$) and at the end it has high variance between $\sim 1$ and $\sim 0.5$.

%%%%%%%%%%%%%%%%%%%%%%%%%%%%%%%%%%%%%%%%%%%%

\section{Conclusions}
\label{sec:conclusions}

In this study, we have proposed a Python framework for learning parameter adaptation policies in metaheuristics. The framework, based on a state-of-the-art RL algorithm (PPO) is of general applicability and can be easily extended to handle various optimizers and parameters thereof.

In the experimentation, we have applied the proposed framework to the learning of the step-size in CMA-ES, and the scale factor and crossover rate in DE. Our experiments demonstrate the efficacy of the learned adaptation policies, especially considering the Best of Run results in the case of DE, in comparison with well-known adaptation policies taken from the literature such as iDE and jDE.

The hybridization of metaheuristics and RL, to which this paper contributes, is becoming growing field of research, and offers the potential to create genuinely adaptive numerical optimization techniques, with the possibility to perform continual learning and incorporate previous knowledge. In this regard, this work can be extended in multiple ways. The most straightforward direction would be to test alternative RL models (different from PPO). Moreover, while in this study we focused on real-valued optimization, in principle the proposed system could be extended to handle parameter adaptation also for solving combinatorial problems. Furthermore, it will be important to test the proposed framework in real-world applications, and include in the comparative analysis other state-of-the-art optimizers. Moreover, it would be interesting to investigate alternative observation spaces and reward functions. Another option would be to extend the framework to learn the choice of operators and algorithms (in an algorithm portfolio scenario), rather than their parameters.

%%%%%%%%%%%%%%%%%%%%%%%%%%%%%%%%%%%%%%%%%%%%

\begin{acks}
We thank Alessandro Cacco for a preliminary implementation of the framework used in the experiments reported in this study.
\end{acks}

%%%%%%%%%%%%%%%%%%%%%%%%%%%%%%%%%%%%%%%%%%%%

\balance

\bibliographystyle{ACM-Reference-Format}
\bibliography{main,sample-base}

%%%%%%%%%%%%%%%%%%%%%%%%%%%%%%%%%%%%%%%%%%%%
%% If your work has an appendix, this is the place to put it.
%\appendix

\end{document}